\pdfoutput=1

\documentclass[11pt]{article}

\usepackage[final]{emnlp2021}

\usepackage{times}
\usepackage{latexsym}
\usepackage{xcolor,colortbl}

\usepackage[T1]{fontenc}

\usepackage[utf8]{inputenc}

\usepackage{microtype}

\usepackage{multirow}
\usepackage{amsmath}
\usepackage{mathtools}
\usepackage{enumerate}
\usepackage{enumitem}
\usepackage{ctable}

\newenvironment{packed_item}{
\begin{itemize}[topsep=.07cm,leftmargin=.2in]
   \setlength{\itemsep}{1pt}
   \setlength{\parskip}{0pt}
   \setlength{\parsep}{0pt}
}{\end{itemize}}

\newcommand{\citenoun}[1]{{\citet{#1}}}

\title{Enhancing Model Robustness and Fairness with Causality: \\A Regularization Approach}

\author{Zhao Wang \\
  University of Chicago \\
  \texttt{zwang13@uchicago.edu} \\\And
  Kai Shu \\
  Illinois Institute of Technology \\
  \texttt{kshu@iit.edu} \\\And
  Aron Culotta \\
  Tulane University \\
  \texttt{aculotta@tulane.edu}}

\begin{document}
\maketitle
\begin{abstract}

Recent work has raised concerns on the risk of spurious correlations and unintended biases in statistical machine learning models that threaten model robustness and fairness. In this paper, we propose a simple and intuitive regularization approach to integrate causal knowledge during model training and build a robust and fair model by emphasizing causal features and de-emphasizing spurious features. Specifically, we first manually identify causal and spurious features with principles inspired from the counterfactual framework of causal inference. Then, we propose a regularization approach to penalize causal and spurious features separately. By adjusting the strength of the penalty for each type of feature, we build a predictive model that relies more on causal features and less on non-causal features. We conduct experiments to evaluate model robustness and fairness on three datasets with multiple metrics. Empirical results show that the new models built with causal awareness significantly improve model robustness with respect to counterfactual texts and model fairness with respect to sensitive attributes.
\end{abstract}

\section{Introduction}\label{sec.intro}

Modern machine learning models are becoming increasingly successful and are widely used in high-stake applications such as filtering loan applicants~\cite{Hassani2020SocietalBR}, determining school admissions~\cite{10.1145/3375627.3375871}, and medical diagnosis~\cite{Ahuja2019TheIO}, etc. However, a big challenge for statistical machine learning is that the models are data-driven and usually built on statistical correlations that are sometimes spurious. For example, a sentiment classifier trained on IMDB movie reviews predicts {\it ``the film directed by Spielberg is incredibly interesting''} as positive. While the prediction is correct, the model takes {\it ``Spielberg''} as a highly predictive feature, that is learned to be strongly correlated with positive sentiment during model training. This is a spurious correlation since an annotator won't label a review as positive just because it mentions {\it ``Spielberg''}. A classifier built on this spurious correlation might perform well when the testing data has the same distribution as training data (i.e., positive reviews talking about {\it ``Spielberg''}), however, it is very likely to fail when there's a distribution shift in testing data (i.e., negative reviews talking about {\it ``Spielberg''}). For another example, a classifier trained on school admission data learns the {\it ``gender''} attribute to be predictive of admission status. The classifier learns this correlation because the {\it ``gender''} attribute happens to correlate with admission in the training data that might contain discrimination and societal bias. Such a biased classifier will raise issues of discrimination when deployed in the real world. In the above examples, the spurious correlations are built in the model because those features happen to correlate with a specific class in the training data, but model designers do not want such features to carry predictive power in the model because that will make the model suffer from robustness (e.g., fail on different distributions), fairness (e.g., bias towards specific groups), and trustworthiness (e.g., not convincing when explaining model performance).


\begin{table}
\centering \small
\begin{tabular}{l} 
 {\it the film directed by \colorbox{brown!30}{Spielberg} is incredibly \colorbox{brown!30}{interesting}.} \\
 ~~~~~~~~~~~~~~~~~~~~~~~~~~~~~~~~~~ 0.712 ~~~~~~~~~~~~~~~~~~~~~~~~~ 0.576 \\
 ~~~~~~~~~~~~~~~~~~~~~~~~~~~~~~~~~~ 0.290 ~~~~~~~~~~~~~~~~~~~~~~~~~ 0.874 \\

\end{tabular}
\caption{The motivating example: the traditional classifier (1st row) learns spurious correlation between {\it Spielberg} and positive sentiment and assigns a large weight for the spurious feature. Our regularization approach (2nd row) encourages the model to assign a smaller weight for the spurious feature and a larger weight for the causal feature ({\it interesting}).}
\label{tab.motivation}
\end{table}

Existing works have investigated multiple ways to deal with spurious correlations and improve model robustness and fairness, such as feature selection~\cite{paul-2017-feature, wang-culotta-2020-identifying}, data augmentation~\cite{kaushik2019learning, Wang2021RobustnessTS}, instance re-weighting~\cite{Zhang2020DemographicsSN},  counterfactual logit pairing~\cite{garg2019counterfactual}, control for confounders~\cite{Virgile2018confounding}, apply constraints to design fair and robust objective functions~\cite{Dwork2012FairnessTA, Zhao2021YouCS}. Recent works also explore causal inference for robust and fair models, such as leveraging human commonsense of causal reasoning~\cite{Srivastava2020RobustnessTS}, adapting text embeddings for causal inference~\cite{pmlr-v124-veitch20a}, and providing causal views on robustness of neural networks~\cite{zhang2020a}. While effective, most works only deal with either fairness or robustness, some works require significant human efforts, some have trade-offs between model robustness and overall performance on test set, and some are complex to deploy in real-world applications.


In this paper, we propose a simple and intuitive regularization approach to integrate causal knowledge in model training. We assign large regularization penalties on spurious features and small penalties on causal features (i.e., features that cause a sample to get a label). By doing so, we encourage the model to pay less attention to spurious features and more to causal features. Table~\ref{tab.motivation} demonstrates a motivating example, where a traditional sentiment classifier (such as Logistic Regression) assigns a large weight for the spurious feature {\it ``Spielberg (0.874)''} and a small weight for the causal feature {\it ``interesting (0.576)''}. The new classifier built with our regularization approach decreases the weight of the spurious feature and increases the weight of the causal feature. 



Specifically, we first manually identify causal and spurious features based on criteria motivated by the counterfactual framework of causal inference. Then, we incorporate regularization components to add different penalties for different types of features. By adjusting the strength of the penalty for each type of feature and optimizing the customized loss function, we build a model that assigns larger weights to causal features and smaller weights to non-causal features. Finally, we carry out experiments to test whether our approach improves model robustness and fairness. For model robustness, we experiment with IMDB movie reviews and Kindle book reviews text data. We evaluate the model performance on both the standard test set as well as a counterfactual test set. Results show that our models significantly improves robustness on the counterfactual test set compared to the traditional models (e.g., 12\% increase in accuracy for the best model). For model fairness, we experiment with the law school admission dataset and evaluate with two metrics (i.e, equal opportunity and demographic parity). Our models show more fair performance than the traditional model (e.g., females and males almost get equal opportunity and demographic parity in the best model). With this framing, the main contributions of this paper are as follows:

    
{\bf Method}: We propose a simple and novel regularization approach to integrate causality during model training. By constructing and optimizing the customized loss function that adjusts the penalty for causal and spurious features, we build predictive models that pay more attention to causal features and less to non-causal features.
    
{\bf Evaluation}: We evaluate our proposed models on both robustness (with high-dimensional text data) and fairness (with low-dimensional tabular data). Our models outperform multiple baselines in terms of robustness to counterfactual texts and fairness to sensitive attributes, without sacrificing model performance on test set.

\section{Related Work}\label{sec.related}
As the main causes for fairness and robustness issues in ML, biases and spurious correlations can be introduced from multiple sources such as culture, data, and algorithm perspectives. Examples include societal discrimination and bias~\cite{10.1145/3457607, Yapo2018EthicalIO}, data imbalance, sampling bias, data poisoning~\cite{Zhao2018GenderBI, kiritchenko-mohammad-2018-examining, chen2017targeted}; covariate shift and confounders~\cite{Virgile2018confounding}, over-parameterization~\cite{sagawa2020investigation,sagawa2020distributionally}.

{\bf Fairness in ML} can be summarized as individual fairness~\cite{Lahoti2019OperationalizingIF}, group fairness~\cite{Hardt2016EqualityOO}, and max-min fairness~\cite{Lahoti2020FairnessWD}. Related research typically falls into three categories based on the stage at which fairness is applied: (1) pre-processing: modify training data and revise sensitive attributes~\cite{Kamiran2011DataPT}, generate balanced data, increase data diversity~\cite{Xu2018FairGANFG}, instance reweighting~\cite{Zhang2020DemographicsSN}; (2) in-processing: apply fairness constraints (regularizations) to design objective functions~\cite{Dwork2012FairnessTA, Zhao2021YouCS}, adversarially learn fair representations~\cite{Beutel2017DataDA}; (3) post-processing: use active learning to collect feedbacks and correct predictions~\cite{zaidan-etal-2007-using}. Here we focus on group fairness with respect to sensitive attributes.

{\bf Robustness in ML-based NLP models}:  \citenoun{Ribeiro2020BeyondAB} propose CheckList to identify critical model failures with diverse types of test cases; \citenoun{garg2019counterfactual} introduce hard ablation, blindness, and counterfactual logit pairing to improve counterfactual token fairness. \citenoun{kaushik2019learning} do counterfactual data augmentmentation to improve model robustness. \citenoun{Lu2020GenderBI} create gender-balanced dataset to learn embeddings that mitigate gender stereotypes. Others explore robust optimization, adversarial training, and domain adaptation methods to improve model robustness~\cite{Namkoong2016StochasticGM,Beutel2017DataDA,BenDavid2006AnalysisOR}.

Recent research draws connections between robustness and causal inference in text. \citenoun{keith2020text} and \citenoun{wooddoughty2018challenges} provide detailed overviews. The works closely related to this paper include: achieving robustness by leveraging human commonsense and counterfactual reasoning~\cite{Srivastava2020RobustnessTS}, adopting active learning and feedback mechanism to highlight rationales~\cite{zaidan-etal-2007-using}, automatically generating counterfactuals with causal words~\cite{Wang2021RobustnessTS}, applying causal inference for feature selection in text classification~\cite{paul-2017-feature, wang-culotta-2020-identifying}, controlling for confounders~\cite{Virgile2018confounding}. Additional works integrate causal inference in text representation and deep learning, such as causal embedding~\cite{pmlr-v124-veitch20a} and causal view on robustness of neural networks~\cite{zhang2020a}.

Our proposed approach is inherently different from the aforementioned approaches: (1) We propose a general regularization approach to achieve robustness and fairness with causality. While previous works have shown the ability to improve model fairness and robustness, most of them only deal with either fairness (to sensitive attributes) or robustness (to counterfactual examples); (2) Our proposed regularization approach is effective (with high prediction performance), efficient (requires little human annotation), and explainable (larger weights for causal features).

\section{Problem Definition}\label{sec.definition}

Our primary task is given input $x$, where $x$ is represented by a sequence of features $[x_1,x_2,...,x_n]$, to predict an outcome y. We consider the case of a classifier $f$ parameterized by $w$ that produces a prediction ${\hat y} = f_w(x)$. The parameters $w$ are learned during model training by optimizing (minimizing) the loss denoted by the error between $y$ and ${\hat y}$.

To clearly illustrate the problem addressed in this paper, we first consider the simple approach of logistic regression classifier $f_w(x) = \frac{1}{1+e^{-\langle x, w \rangle}}$ (\S\ref{sec.LSTM} discusses applying other complex models). We estimate the parameters $w$ on labeled data $\mathcal{X}$ and then examine the (partial) correlations between features and labels by model coefficients (i.e., feature weights). 

The goal is to train a model that performs well on test set and generalizes well on data from different distributions, so that the model maintains robustness to counterfactual data as well as fairness to sensitive attributes (e.g., gender, race). 

\subsection{Robustness to Counterfactuals}\label{sec.robustness}

We measure model robustness on counterfactual samples. If a model is built on spurious correlations, its performance might not drop on a test set that has the same distribution as the training set, but the performance will drop sharply on the counterfactual set, which is created by editing samples from the test set towards a counterfactual label.

Taking text classification as an example, for a piece of text $T$, the corresponding counterfactual sample is created by editing $T$ with minimum changes towards a counterfactual label~\cite{kaushik2019learning}. For example, for the positive review {\it ``the film directed by Spielberg is incredibly {\underline{interesting}}''}, the counterfactual negative review is {\it ``the film directed by Spielberg is incredibly {\underline{boring}}''}. If the classifier is built on spurious correlations such as the correlation between {\it ``Spielberg''} and positive sentiment, it is very likely to make wrong predictions for the counterfactual sentences (e.g., negative reviews talking about {\it ``Spielberg''}), which makes the classifier suffer from robustness on counterfactual data. However, if the classifier is built on causal associations such as the association between {\it ``interesting''} and positive sentiment or {\it ``boring''} and negative sentiment, the model will make right predictions for the counterfactual sentences and for datasets that have different distributions.


\subsection{Fairness to Sensitive Attributes}\label{sec.fairness}

Studies have shown that when there are patterns of previous discrimination and societal bias in the training data, the trained model is very likely to inherit the bias~\cite{10.1145/3457607, Yapo2018EthicalIO}. In this paper, we study group fairness with respect to sensitive attributes (e.g., gender, race). The goal is to train a model that ensures members of protected groups in the population (e.g., based on sensitive demographic attributes like gender or race) receive ``fair share of beneficial outcomes'' during model prediction~\cite{Hardt2016EqualityOO}.

Taking the law school admission data as an example, we want to train a classifier that is fair to populations of different identity groups defined by sensitive attributes (e.g., gender, race). If the classifier is built on features such as test scores for the admission decision, the model should make fair predictions for applicants from different identity groups. However, if the classifier is built on biased correlations such as using {\it ``gender''} as a predictive feature for the admission decision, it could potentially harm end-users who identify with those groups and cause ethical concerns. 

In this work, we study group fairness with the low-dimensional law school admission data and it's possible to extend the proposed approach to high-dimensional text data with additional text processing. For example, in the task of toxicity prediction of online comments, we can first categorize the comments into different identity groups according to the sensitive attributes they describe (e.g., race, sexual orientation), and then improve classifier fairness with respect to the protected groups using our regularization approach. One big difference between studying fairness with low-dimensional tabular data (e.g., law school admission data) and high-dimensional text data is that for low-dimensional data, we directly identify the sensitive attributes as spurious features, but for text data, we need to identify keywords that are descriptive of sensitive attributes as spurious features (e.g., ``black'' and ``white'' are reflective of race, ``gay'' and ``straight'' are reflective of sexual orientation). We will take this as our future work.







\section{Methods}\label{sec.methods}

To build a model that is robust to counterfactuals and fair to sensitive attributes, we need to deal with issues of spurious correlations and biases. Our solution is a two-stage process. We first manually identify causal and spurious features based on the criteria motivated by the counterfactual framework of causal inference. We then propose a regularization approach to build a loss function that adds small penalties for causal features and large penalties for spurious features. By doing so, the model is encouraged to rely more on meaningful causal associations and less on spurious correlations.

\subsection{Annotate Causal and Spurious Features}\label{sec.features}

Consider an instance $x$ represented by a sequence of features $[x_1,x_2,...,x_n]$ and having the label $y$. We identify causal and spurious features by using the counterfactual framework of causal inference~\cite{winship1999estimation}: if feature $x_i$ were replaced with some other feature $x_j$, how likely is it that the label $y$ would change? In this paper, we consider short texts (e.g., single sentences) and low-dimensional tabular data as the unit of analysis, and the following criteria is proposed for this type of data.

A {\it causal feature} is a feature $x_i$ that causes the sample $x$ to receive the label $y$. All else being equal, one would expect $x_i$ to be a determining factor in assigning $y$ to sample $x$. 
On the other hand, a {\it spurious feature} is a feature $x_i$ that correlates with the target class $y$ in a specific dataset, but replacing it with another feature $x_j$ would be unlikely to change the instance label.

Taking a movie review sentence as an example, {\it ``the film directed by Spielberg is incredibly \underline{interesting}'' (pos)}, ``interesting'' is a causal feature that is primarily responsible for the positive sentiment and replacing it with another word such as ``boring'' will change the sentiment to be negative. In contrast, ``Spielberg'' is a spurious feature as it does not convey positive sentiment nor cause a review to be positive, and replacing it with another word such as ``Kevin'' will not change the sentiment. Although ``Spielberg'' might correlate with positive sentiment due to its high frequency in positive movie reviews, this statistical correlation does not imply causality.

With the proposed criteria, we manually annotate causal and spurious features in \S\ref{sec:annotate_features}. Table~\ref{tab.example_features} shows examples of those features identified for sentiment classification and school admission tasks.

Prior works have explored methods that automatically identify causal and spurious features. For example, \citenoun{polyjuice:acl21} design Polyjuice as a general-purpose counterfactual generator that allows for control over multiple perturbation types and the generated counterfactuals are application agnostic. \citenoun{wang-culotta-2020-identifying} propose to fit a classifier for auto-prediction of causal and spurious features using information such as word embeddings and individual treatment effects. \citenoun{Wang2021RobustnessTS} propose a closest opposite matching approach for auto-identification of likely causal features. While these methods work well in general, each has its drawbacks. In this paper, we want to focus on the regularization effect and isolate the noises introduced by other procedures, so we manually annotate causal and spurious features as ground truth.





\begin{table}
\centering \small
\begin{tabular}{l|ll}
\hline
 & {\bf Causal} & {\bf Spurious}\\
\hline
\multirow{3}{*}{Sentiment Classification} & interesting & movie \\
 & boring & story  \\
 & wonderful & Spielberg \\
 & inspiring & definitely \\
 & enjoyed & animated \\
\hline
\multirow{2}{*}{School Admission} & LSAT & Gender \\
 & GPA & Race \\
\hline
\end{tabular}
\caption{Examples of causal and spurious features}
\label{tab.example_features}
\end{table}

\subsection{Regularization with Causal and Spurious Features}\label{sec.loss}
Our goal is to train a model that performs well on test data (that has the same distribution as training set) as well as maintains robustness to counterfactuals and fairness to sensitive attributes. Prior works~\cite{kaushik2019learning, Wang2021RobustnessTS} have demonstrated the effectiveness of augmenting training data with counterfactuals to improve model robustness. \citenoun{Wang2021RobustnessTS} conducted empirical analysis to compare the difference between a traditional model and a robust model, finding that the robust model assigns larger weights for causal features and smaller weights for spurious features.

While previous works have explored the effectiveness of adding L2/L1 penalty (regularization) to prevent over-fitting~\cite{Ng2004FeatureSL}, the penalty is applied on all features. In this paper, we design a customized loss function that adds different penalties for different types of features (e.g, causal, spurious, and the remaining). The intuition is that when tuning a classifier, we want it to pay more attention to causal features (small penalty) and less attention to spurious features (large penalty). Below we consider three sets of features: causal ($C$), spurious ($S$), and remain ($R$, i.e., features except for those labeled as causal or spurious). 


Let $J$ be the loss function representing the error/difference between the truth and prediction. To add different penalties for causal, spurious, and remaining features separately, we design the loss function as:

\begin{equation}
\begin{aligned}
    \sum_{x \in X}{J(f(x),y)} +
    \frac{\lambda_c}{|C|}\sum_{c \in C}{w_c}^2 \\
    + \frac{\lambda_s}{|S|}\sum_{s \in S}{w_s}^2 \\ + \frac{\lambda_r}{|R|}\sum_{r \in R}{w_r}^2    
\end{aligned}
\end{equation}
where $\lambda_c$, $\lambda_s$, and $\lambda_r$ refer to the strength of penalty for causal, spurious, and remaining features; $|C|$, $|S|$, and $|R|$ refer to the size of corresponding feature set; $w_c$, $w_s$, and $w_r$ refer to the weights of features from each set. 

There are many variations of this loss function. For example, we can set $\lambda_c = 0$ and $\lambda_r = 0$ to only penalize spurious features. We can also set $\lambda_c = 0$ to only penalize non-causal features (i.e., a combination of spurious and remain features).




The primary advantage of this approach is that it integrates causal knowledge into model training and provides the flexibility to penalize and address the importance of different types of features. By optimizing the proposed loss function, we seek to build classifiers that satisfy two desirable properties: (1) perform well on testing data and (2) pay more attention to causal features and less to spurious/non-causal features, which is the key to make the model robust and fair on data from different distributions.


\section{Experiments}\label{sec.exp}
We conduct experiments with three datasets to evaluate the effectiveness of the proposed method from two aspects\footnote{Code and data available at: \url{https://github.com/tapilab/emnlp-2021-regularization}}: robustness to counterfactuals (evaluate on high-dimensional text data from IMDB movie reviews and Kindle book reviews) and fairness to sensitive attributes (evaluate on low-dimensional tabular data of law school admission).

\subsection{Data}\label{sec.data}

\textbf{IMDB movie reviews:} This dataset is collected and published by~\citenoun{kaushik2019learning}. They first randomly sampled $2.5K$ reviews with balanced class distributions from the original large IMDB movie reviews dataset~\cite{pang2005seeing} and then instruct Amazon Mechanical Turk workers to edit each review with minimum changes towards a counterfactual label. In the final dataset, every review gets a corresponding counterfactual text. We apply the data splitting criteria from the paper and the train/validation/test sets contain 1,707/245/488 samples. The counterfactual test set is created by counterfactually editing samples from the test set.


\textbf{Kindle book reviews:} This dataset\footnote{\url{https://jmcauley.ucsd.edu/data/amazon/}} contains book reviews from the Amazon Kindle store~\cite{He_2016} and each review is labeled as positive or negative based on its rating (ratings \{4,5\} as positive and ratings \{1,2\} as negative). To reduce noisy samples, we limit the reviews to those contain 5 to 40 words and the final dataset is split into train/validation/test sets containing 7,500/2,500/500 samples. For the 500 test samples, we manually edit each sample with minimum changes towards a counterfactual label (i.e., the same criteria used in~\cite{kaushik2019learning} to edit counterfactuals for IMDB reviews above).

\textbf{Law school admission data:} This dataset contains admission data of 25 US law schools over 2005-2007\footnote{\url{http://www.seaphe.org/databases.php}}. Each sample is an admission record for a student and is represented by seven features: LSAT, GPA, resident, admission year, gender, race, URM (under-represented minority), with a binary label indicating the admission status. The dataset is split into train/validation/test sets containing 50\%, 20\%, and 30\% samples.

For the IMDB and Kindle datasets, each sample $x$ is a piece of text, represented by a binary, bag-of-words representation $x = [x_1, x_2, ..., x_v]$, where $v$ is the vocabulary size and $x_i$ is a binary value indicating the appearance of a word in the text. Compared with the high-dimensional text data, the law school admission data is low-dimension with each sample represented by seven numerical or categorical features. We process this dataset by first applying one-hot encoding for categorical features and then normalizing numerical features into range 0-1. In the experiments below, we show that our method works well on both high- and low-dimensional data.


\begin{table}
\centering \small
\begin{tabular}{l|lll}
\hline
\textbf{} & \textbf{IMDB} & \textbf{Kindle} & \textbf{Admission} \\
\hline
\#Instances& 2.5k & 10.5k & 65.3k \\
\#Features & 2,388 & 2,041 & 7 \\
\#Causal(top) & 65 & 76 & 2 \\
\#Spurious(top) & 166 & 118 & 1 \\
\#Causal(vocabulary) & 362 & 293 & 2 \\
\hline
\end{tabular}
\caption{Data summary}
\label{tab.data}
\end{table}

\subsection{Causal and Spurious Features}\label{sec:annotate_features}

With the criteria proposed in~\ref{sec.features}, we ask two student annotators to manually label causal and spurious features from each dataset. Specifically, for each dataset, we first train an initial classifier and extract top coefficient features (e.g. coefficient magnitude > 1). For example, for logistic regression model, we would extract features with high magnitude coefficients. For more complex models, other transparency algorithms may be used~\cite{martens2014explaining}. 

For the IMDB and Kindle datasets, the annotators identified two sets of causal and spurious features: the first set is annotated from the top coefficient features and the second set is annotated from the whole vocabulary (the first set is a subset of the second set). While there is some subjectivity during annotation, we did a round of training to resolve disagreements prior to annotation and the final agreement was generally high for the experimental datasets (on average 96\% raw agreement by fraction of labels that agree).

For the law school admission dataset, there are only seven features and the annotators directly identify {\it ``LSAT''} and {\it ``GPA''} as causal features and {\it ``gender''} as a spurious feature (i.e., sensitive attribute). There are other  spurious features such as {\it ``race''} and {\it``URM''}, but for this paper, we only focus on model fairness with respect to gender (to have a direct comparison with previous work), so we identify {\it ``gender''} as a spurious feature and take other potential spurious features as ``remaining'' features.

Table~\ref{tab.data} summarizes the number of identified causal and spurious features for each dataset.

 

\begin{table*}
    \centering \small
    \begin{tabular}{l| l l l | l l l} 
    \hline
      & \multicolumn{3}{c|}{{\bf IMDB}} & \multicolumn{3}{c}{{\bf Kindle}} \\
        
         {\bf Method} &  & {\bf Test} & {\bf CTF} &  & {\bf Test} & {\bf CTF} \\
         \hline
         Feature selection &  & 0.835 $\pm$ 0.005 & \cellcolor{brown!30}{0.633 $\pm$ 0.012} &  & 0.850 $\pm$ 0.002 & 0.510 $\pm$ 0.017 \\ 
         Data augmentation  &  & 0.818 & \cellcolor{brown!30}{0.869} & & 0.752 & \cellcolor{brown!30}{0.720} \\ 
         L2 with BOW &  & 0.845 $\pm$ 0.005 & 0.567 $\pm$ 0.009 &  & 0.853 $\pm$ 0.027 & 0.412 $\pm$ 0.048 \\
         L2 with Glove &  & 0.824 $\pm$ 0.002 & 0.619 $\pm$ 0.006 &  & 0.820 $\pm$ 0.004 & \cellcolor{brown!30}{0.553 $\pm$ 0.027} \\
         \specialrule{.1em}{.1em}{.1em}
         

         {\bf Regularization} ({\it top})& {\bf $\lambda_c,\lambda_s,\lambda_r$} &  &  & {\bf $\lambda_c,\lambda_s,\lambda_r$} &  &  \\
         \hline
         
         causal+spurious & (0,10,0) & 0.836 $\pm$ 0.006 & 0.632 $\pm$ 0.005 & (0,10,0) & 0.857 $\pm$ 0.007 & 0.545 $\pm$ 0.013\\
         causal+non-causal & (0,100,100) & 0.821 $\pm$ 0.011 & \cellcolor{brown!30}{0.658 $\pm$ 0.01} & (0,10,10) & 0.843 $\pm$ 0.003 & 0.557 $\pm$ 0.003 \\
         causal+spurious+remain & (0,100,0.1) & 0.841 $\pm$ 0.007 & 0.635 $\pm$ 0.006 & (0,100,1) & 0.847 $\pm$ 0.003 & \cellcolor{brown!30}{0.571 $\pm$ 0.008} \\
         \hline
          
         {\bf Regularization} ({\it vocab}) & & & & & & \\
         causal+spurious & (0,10,0) & 0.838 $\pm$ 0.004 & 0.634 $\pm$ 0.007 & (0,100,0) & 0.857 $\pm$ 0.003 & 0.545 $\pm$ 0.01\\
         causal+non-causal & (0,100,100) & 0.843 $\pm$ 0.005 & \cellcolor{brown!30}{0.700 $\pm$ 0.007} & (0,10,10) & 0.855 $\pm$ 0.005 & \cellcolor{brown!30}{0.641 $\pm$ 0.006} \\
          causal+spurious+remain & (0.01,100,100) & 0.840 $\pm$ 0.004 & \cellcolor{brown!30}{0.697 $\pm$ 0.008} & (0.001,1,10) & 0.848 $\pm$ 0.006 & \cellcolor{brown!30}{0.636 $\pm$ 0.003} \\
         \hline
         
    \end{tabular}
    \caption{Robustness evaluation for tuned classifiers. Accuracy reported for test and Counterfactual (CTF) data from IMDB and Kindle dataset. Results in 1st section are baseline and state-of-the-art methods, 2nd section shows regularization approach using causal features identified from top coefficient terms and 3rd section shows regularization approach using causal features identified from the whole vocabulary.}
    \label{tab.CTF_robust}
\end{table*}

\subsection{Experimental Settings}\label{sec.baseline}

Among the numerous works focused on improving model robustness to counterfactuals, the ones most closely related to ours include: augmenting training data with counterfactuals to train a robust classifier~\cite{kaushik2019learning, Wang2021RobustnessTS}; a logit pairing strategy to penalize the norm of difference in logits of training examples and their counterfactuals~\cite{garg2019counterfactual}; causal inference for feature selection~\cite{paul-2017-feature}; a feature selection strategy to remove spurious features and improve model robustness with respect to the worst-case accuracy~\cite{wang-culotta-2020-identifying}.

We compare our proposed regularization approach with the representative and state-of-the-art prior works:
\begin{packed_item}
    \item {\bf Feature Selection}~\cite{wang-culotta-2020-identifying}: we explore whether directly removing spurious features is more effective than penalizing spurious features with regularization approach.
    
    \item {\bf Data Augmentation}~\cite{Wang2021RobustnessTS}: Augmenting original training data with corresponding counterfactual data (either automatically generated or manually edited). This method combines counterfactuals as part of the training data and changes data distribution and data size. It usually requires significant human efforts to generate/edit counterfactuals.
\end{packed_item}

For our proposed regularization approach, we experiment with the LogisticRegression classifier and explore the following settings:
\begin{packed_item}
    \item {\bf L2-regularization}: as representative of the traditional regularization approach.
    
    \item {\bf Feature representation for text}: we explore Bag-Of-Words (BOW) and pre-trained Glove embedding~\cite{pennington2014glove} representations to check whether complex representations help improve robustness.
    
    \item {\bf The number of causal and spurious features}: we experiment with  two sets of causal and spurious features: (a) a small set identified from top coefficient features and (b) a full set identified from the whole vocabulary. The goal is to check whether providing more causal features improves model robustness.
    
\end{packed_item}

{\bf Hyper-parameter tuning}: we use BinaryCrossEntropy as the basic loss function in all models and implement the regularization approach by customizing the loss function with penalties for causal, spurious, and remaining features as discussed in~\S\ref{sec.loss}. The penalty strengths (i.e., $\lambda_c, \lambda_s, \lambda_r$) are explored in the range [0, 0.0001, 0.001, 0.01, 0.1, 1, 10, 100, 1000] with the constraint that $l_s \ge l_r > l_c$.
We use Adam optimizer and set the initial learning rate to 0.001, and adjust the learning rate according to different $\lambda$ values to avoid overshooting. We use early stopping criteria on validation data to determine the epoch to stop tuning (i.e., stop tuning if the validation loss hasn't decreased for 10 epochs). Among all the tuned models, we select the best model based on its performance on the counterfactual validation data.

After selecting the best model, we evaluate its performance on the test set and counterfactual set. We repeat the training process 10 times with different random seeds and report the model performance and corresponding hyper-parameter settings in Table~\ref{tab.CTF_robust}. All experiments are conducted on a 64-bit machine with Nvidia GPU (Tesla V100, 1246MHz, 16 GB memory).



\subsection{Robustness Evaluation}\label{sec.exp_robust}
We evaluate model robustness on the counterfactual texts from IMDB and Kindle datasets. We train different models by optimizing the customized loss functions suggested in~\S\ref{sec.loss} and evaluate model performance on the test set and counterfactual set. The best model parameter setting and performance is reported in Table~\ref{tab.CTF_robust}.

Results show that: (1) our proposed regularization approach outperforms feature selection on the counterfactual test set without sacrificing performance on the test set; data augmentation outperforms our regularization approach on the counterfactual set but sacrifices performance on the test set, which is expected because data augmentation approach increased training data size and changed training data distribution. And data augmentation usually requires more human efforts than our regularization approach. (2) separately penalizing causal, spurious, and remaining features is more effective than L2 regularization in terms of model robustness on counterfactual set. The best performance is achieved by assigning no/small penalty for causal features and large penalties for spurious and remaining features; (3) the more causal features provided, the better robustness on counterfactual set. For IMDB, the best model gives 9\% improvement with top causal features and 13.3\% with all causal features. For Kindle, the best model gives 15.9\% improvement with top causal features and 22.9\% with all causal features; (4) complex text representation (e.g., embedding) has limited improvements on model robustness.

Note that the results and parameter settings in Table~\ref{tab.CTF_robust} are based on the exploration of penalty strengths (i.e.,$\lambda_c, \lambda_s, \lambda_r$) in the range of [0, 0.0001, 0.001, 0.01, 0.1, 1, 10, 100, 1000]. The best parameter settings might change for different datasets and larger search space. According to our exploration about the effect of hyperparameter $\lambda$ on model performance in ~\S\ref{sec:lambda}, we recommend to set $\lambda_c <= 0.01$, $\lambda_s >= 1$, and $\lambda_r >= 1$ for experiments on new datasets that don't have counterfactual validation set to decide the best hyperparameter setting. As a reference, when we fix $\lambda_c = 0$, $\lambda_s = 100$, and $\lambda_r = 10$ for all datasets, the accuracy is within 97\% of the best found setting.

\begin{figure}[t]
	\centering
	\includegraphics[width=0.5\textwidth]{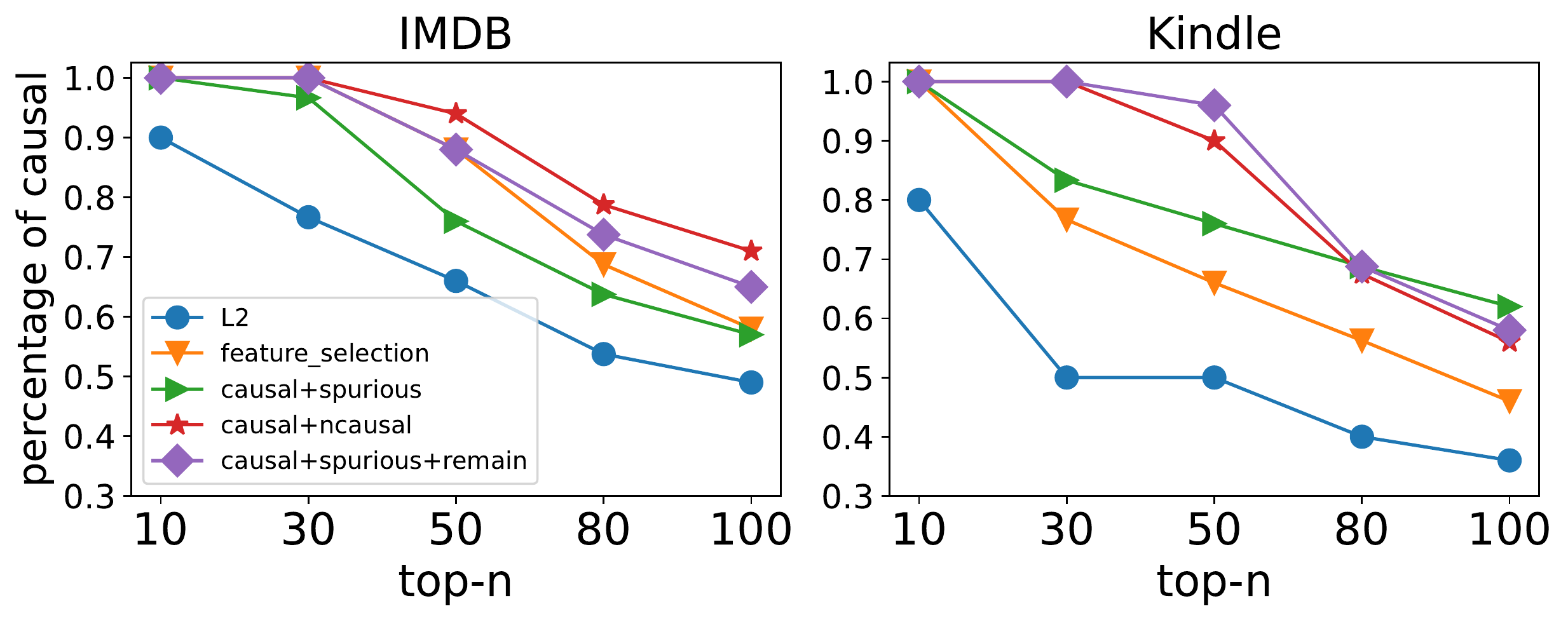}
	\caption{Percentage of causal features among the top-n features of each model. The three models tuned with penalties for causal, spurious, and remain features refer to the models shown in the last three lines in Table~\ref{tab.CTF_robust} ({\it best viewed with color}).}
	\label{fig.p_causal}
\end{figure}

We also show the percentage of causal features among top-n large coefficient features of each model in Fig~\ref{fig.p_causal}. Top features are features that are most important and predictive of the classes, and a high percentage of causal features means that the model relies more on causal features and less on others. We can see that models tuned from our proposed loss functions have much higher percentage of causal features compared with the baselines (L2 and feature selection). The model that only considers causal and spurious features has a relative lower percentage compared with the other two models that consider causal, spurious, and remaining features. Taking the model that separately penalizes causal and non-causal features for an example (IMDB), all features in the top-10 and top-30 are causal features, and about 93\% of top-50 and 70\% of top-100 are causal features. The difference is even bigger for the Kindle dataset. However, for the baseline models, the percentage is much lower. The high percentage of causal features indicates that our regularization approach successfully makes the model rely more on causal features.

	
\begin{table*}
\centering \small
\begin{tabular}{llllll}
\hline
{\bf Penalty} & {\bf $\lambda_c,\lambda_s,\lambda_r$} & accuracy & f1 & {\bf $\Delta_{EO}$} & {\bf $\Delta_{DP}$}\\
\hline
L2 (baseline) & N/A & 0.796 $\pm$ 0.001 & 0.650 $\pm$ 0.012 & 0.102 $\pm$ 0.003 & 0.028 $\pm$ 0.001 \\
\hline
causal+spurious & (0.0001,10,0) & 0.795 $\pm$ 0.003 & 0.654 $\pm$ 0.011 & \cellcolor{brown!30}{0.023 $\pm$ 0.005} & \cellcolor{brown!30}{0.007 $\pm$ 0.001} \\
causal + non-causal & (0.0001,0.001,0.001) & 0.796 $\pm$ 0.001 & 0.655 $\pm$ 0.012 & 0.100 $\pm$ 0.006 & 0.027 $\pm$ 0.001 \\
causal+spurious+remain & (0.0001,0.1,0.01) & 0.794 $\pm$ 0.003 & 0.646 $\pm$ 0.01 & \cellcolor{brown!30}{0.028 $\pm$ 0.005} & \cellcolor{brown!30}{0.004 $\pm$ 0.002} \\
\hline
FairRF~\cite{Zhao2021YouCS} & N/A & 0.796 $\pm$ 0.002 & N/A & 0.023 $\pm$ 0.008 & 0.007 $\pm$ 0.004\\
\hline
\end{tabular}
\caption{Tuned classifier performance on law school admission test set and fairness evaluation}
\label{tab.law_fairness}
\end{table*}

\subsection{Fairness Evaluation}\label{sec.exp_fair}
We experiment with law school admission dataset and ask the question:``does a model trained on this dataset makes fair admission predictions for female and male students?'' We apply the same baseline and training procedure described in~\S\ref{sec.exp_robust} with {\it ``LSAT''} and {\it ``GPA''} as causal features and {\it ``gender''} as spurious feature (i.e., sensitive attribute). Following existing work on fairness evaluation~\cite{Zhao2021YouCS}, we adopt two widely used evaluation metrics, equal opportunity and demographic parity.

{\bf Equal opportunity} evaluates whether a classifier gives similar results for different sensitive groups in the positive class~\cite{Sahil2018, Hardt2016EqualityOO, Yan2020FairCB}. We calculate equal opportunity difference $\Delta_{EO}$ to measure whether a classifier is fair for gender attribute:
\begin{equation}
\begin{aligned}
\Delta_{EO} = |Pr\{\hat{y} = 1| S = i,y = 1\}\\ - Pr\{\hat{y} = 1| S = j,y = 1\}|
\end{aligned}
\end{equation}
where $S$ is sensitive attribute, $i$ and $j$ are sensitive attribute values defined by $S$ (e.g, male and female defined by gender), and $y$ is the truth label.

In our experiment with law school admission data, if a model is fair, then the probability of an applicant being predicted to be admitted ($\hat{y}=1$) should be the same for both male ($S=i$) and female ($S=j$) applicants that are actually being admitted ($y=1$).

{\bf Demographic Parity} measures non- discriminative performance of a classifier. We calculate demographic parity difference $\Delta_{DP}$ to check whether a classifier is fair for different sensitive groups:
\begin{equation}
\begin{aligned}
\Delta_{DP} = Pr\{\hat{y} = 1 | S = i\}\\ -  Pr\{\hat{y} = 1 | S = j\}
\end{aligned}
\end{equation}



We report the evaluation results from the best model and corresponding model parameters in Table~\ref{tab.law_fairness}. Results show that: (1) the model trained with our regularization approach is more fair with respect to the gender attribute and has comparable performance with previous work using complex models~\cite{Zhao2021YouCS}; (2) model fairness is achieved by assigning small/no penalties for causal features and large penalties for non-causal features, which is similar as achieving robustness. 

\begin{figure}[t]
	\centering
	\includegraphics[width=0.5\textwidth]{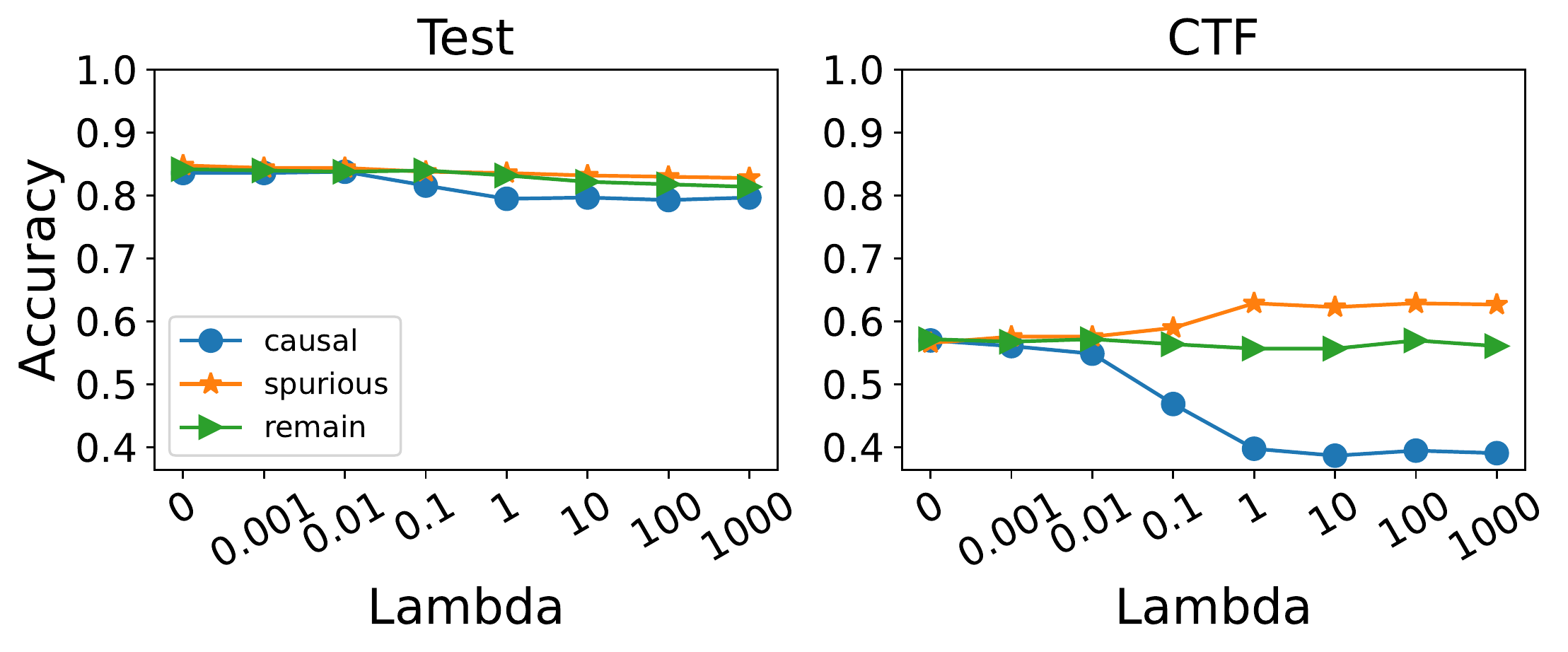}
	\caption{Model performance change with lambda (IMDB) ({\it best viewed with color}).}
	\label{fig.lambda}
\end{figure}

\section{Discussion}

\subsection{The Effect of Hyperparameter $\lambda$}\label{sec:lambda}
To understand the effect of $\lambda_c$, $\lambda_s$, and $\lambda_r$ on model performance, we conduct experiments to first initialize $\lambda$ to 0, and then change the value of each $\lambda$ (Fig~\ref{fig.lambda}). We observe that: (1) model performance on test set is relatively stable with different $\lambda$ parameters; (2) model performance on counterfactual set is mainly affected by $\lambda_c$. By increasing $\lambda_c$, we assign larger penalty on causal features and the model performance drops. In contrast, by increasing $\lambda_s$, we assign larger penalty on spurious features and the model performance increases. Penalizing causal features affects model performance more than penalizing spurious features, and penalizing remaining features has the least effect.





\subsection{Extension in Deep Learning Framework}\label{sec.LSTM}
In the experiments, we use LogisticRegression as our basic classifier because it's intuitive to explain feature weights and penalties. But our main idea is to integrate causality in model training by adding small penalties to causal features and large penalties to non-causal features, and the loss function is not limited to any specific model, so we do think that our approach can be extended to more complex models. For example, possible ways to apply the idea to LSTM classifiers include: (1) assign different attentions for causal, spurious, and remaining features in the input; (2) specify different probabilities to mask out or drop out spurious and causal feature. We leave this part as our future work.





\section{Conclusions}
In this paper, we have introduced the idea of integrating causal knowledge through a regularization approach. We implement this by first identifying causal and spurious features and then assigning different penalties for causal and non-causal features in the loss function. The resulting model assigns larger weights for causal features and smaller weights for spurious features. Experiments have demonstrated that the tuned model is effective to improve robustness and fairness and it works well for both low- and high- dimensional data. The proposed idea is simple to implement, intuitive to explain, has the potential to be easily deployed to other frameworks and various tasks on model robustness and fairness. For future works, we will design automatic methods to identify causal and spurious features and extend the regularization approach to complex deep learning frameworks.

\section*{Acknowledgments}
This  research  was  funded  in  part  by  the  National  Science Foundation under grant \#1618244. We would also like to thank the anonymous reviewers for useful feedbacks.

\bibliography{anthology, reference}
\bibliographystyle{acl_natbib}


\end{document}